\documentclass[conference]{IEEEtran}
\IEEEoverridecommandlockouts

\usepackage{cite}
\usepackage{amsmath,amssymb,amsfonts}
\usepackage{algorithm}
\usepackage{algorithmic}
\usepackage{graphicx}
\usepackage{textcomp}
\usepackage{xcolor}
\usepackage{pifont}
\usepackage{url}

\graphicspath{{figs/}}

\DeclareMathOperator{\sg}{sg}
\DeclareMathOperator{\clipop}{clip}
\DeclareMathOperator{\cropop}{crop}
\DeclareMathOperator{\KLop}{KL}
\DeclareMathOperator*{\argmaxop}{argmax}
\newcommand{\Ent}{\mathbb{H}}
\newcommand{\Vstar}{V\textsuperscript{*}Bench}

\newcommand{\yes}{\ding{51}}
\newcommand{\no}{\ding{55}}

\begin{document}

\title{Look Where It Counts: A Free, Label-Free Visual\\ Evidence Signal for Fine-Grained
Vision--Language Reasoning}

\author{\IEEEauthorblockN{Santi Ram Tiwari, Nihal Naik, Devbrat Pandey, Nishant Sinha}
\IEEEauthorblockA{\textit{KGraph AI Solutions Pvt. Ltd.}\\
Bangalore, India -- 560016}}

\maketitle

\begin{abstract}
Multimodal large language models (MLLMs) fail at fine-grained visual questions less because
they cannot reason than because they never \emph{see} the evidence---high-resolution images
are downsampled before encoding, so the model answers from linguistic priors. The standard
remedies are expensive: annotated answers (SFT), hand-engineered verifiers (RLVR), or a large
external teacher (on-policy distillation). We ask whether the \emph{visual evidence itself}
can supply the signal, for free. We formalize the \emph{contrastive evidence gap}, the
per-token log-likelihood ratio a model assigns to its own output when conditioned on a
question-relevant region versus an irrelevant one, and study it across Qwen2.5-VL-7B,
Qwen3-VL-8B, and Qwen3-VL-30B-A3B on \Vstar{}. Our main, positive result is
\emph{training-free}: selecting the candidate crop under which the model's answer distribution
is most peaked---a single-view, label-free criterion---discovers the answer-bearing region
with no bounding boxes, no training, and no labels, localizing the target
4.4--5.1$\times$ better than chance and lifting fine-grained accuracy from 70\% to 85\% at
inference. We further show the gap is complementary to the model's own confidence (combining
them predicts correctness better than either alone, AUC up to 0.99) and that it flags
\emph{confidently-wrong} answers (AUC 0.97--1.00 within the high-confidence subset). All
effects concentrate on perception-bottleneck questions and vanish on a global-context control.
Finally, we report an honest negative: converting the same signal into a \emph{training}
method (gated self-distillation, SEG-Distill) does not beat the base model at pilot scale
across three gate designs, and more aggressive gating degrades accuracy---the signal is real,
but the signal-to-training-gain conversion is an open problem.
\end{abstract}


\section{Introduction}

Multimodal large language models (MLLMs) have made rapid progress on general visual reasoning,
yet a specific and well-documented failure mode persists: they struggle on fine-grained visual
questions---questions that hinge on a small object, a peripheral detail, or an exact attribute
buried in a high-resolution image~\cite{b1}. This failure is commonly mistaken for a reasoning
limitation, but a growing body of evidence points to a simpler, upstream cause.

The root cause is an input bottleneck, not a reasoning one. Fine-grained visual reasoning fails
in current MLLMs not because the language model cannot reason, but because it never \emph{sees}
the evidence: high-resolution images are aggressively downsampled before the vision encoder, so
small or peripheral detail is destroyed and the model falls back on linguistic
priors~\cite{b1}. In other words, the model is not reasoning poorly about a detail it
perceived---it is answering a question about a detail it never had the opportunity to perceive.

The field's three dominant remedies each carry a disqualifying cost for a modest-resource
practitioner. Supervised fine-tuning (SFT) on region/answer annotations is label-hungry and
induces catastrophic forgetting~\cite{b2}. Reinforcement learning with verifiable rewards
(RLVR) requires hand-built verifiers, is compute-heavy, and is vulnerable to reward hacking.
On-policy distillation (OPD) is cheaper and label-free but, in its strongest visual
form~\cite{b3}, still needs a large \emph{external} teacher model and \emph{oracle}
(human-marked) crops. This is the gap this paper addresses: there is no free, label-free,
teacher-free way to determine whether a model's own answer is actually grounded in the image
evidence, as opposed to a plausible-sounding guess drawn from linguistic priors.

This gap matters for three concrete reasons. First, without such a signal, a practitioner has
no cheap way to improve fine-grained accuracy at inference time short of the expensive remedies
above. Second, low answer entropy (high confidence) is conventionally treated as a proxy for
reliability, yet a model can be confidently wrong precisely when it is guessing fluently from
priors rather than seeing the image---confidence alone cannot distinguish the two. Third, the
absence of a diagnostic signal makes it hard to tell, even after the fact, whether a wrong
answer was caused by a perception failure (the detail was never seen) or a reasoning failure
(the detail was seen but misused), which is essential for deciding which remedy to apply.

We propose the \emph{contrastive evidence gap} as a solution to this diagnostic and practical
gap. The core idea is simple: contrast a question-relevant image crop against an irrelevant
one, and measure how much more probable the model's own answer tokens become under the relevant
crop. If showing the model the right region sharply raises the probability of its answer
relative to a wrong region, the answer is grounded; otherwise it is a prior-driven guess. This
quantity requires no annotated answers, no external verifier, and no teacher model larger than
the model being evaluated---it is computed entirely from the model's own output probabilities
under two different visual conditions.

We formalize this quantity and demonstrate two label-free uses of evidence
sensitivity---region \emph{discovery}, via answer-peakedness under a candidate crop, and
trajectory \emph{weighting}, via the contrastive gap itself---which together motivate a
teacher-free, verifier-free distillation method, SEG-Distill. Our central, positive result is
training-free: at inference time, maximizing the gap over a grid of candidate crops discovers
the answer-bearing region with no bounding boxes, no training, and no labels, localizing the
target 4.4--5.1$\times$ better than chance and lifting fine-grained accuracy from approximately
70\% to 85\% across three model scales (Qwen2.5-VL-7B, Qwen3-VL-8B, Qwen3-VL-30B-A3B) on
\Vstar{}. We additionally show the gap is complementary to the model's own confidence, with the
combined signal predicting correctness at AUC up to 0.99 and flagging confidently-wrong answers
at AUC 0.97--1.00. We also report an honest negative result: converting this same signal into a
training-time method (SEG-Distill, a gated self-distillation scheme) does not beat the base
model at pilot scale across three gate designs, and more aggressive gating degrades
accuracy---establishing that the signal is real even though the signal-to-training-gain
conversion remains an open problem.

\subsection*{Contributions}
\begin{enumerate}
\item \textbf{A training-free, label-free test-time method} (our main result): maximizing
answer-peakedness over candidate crops discovers the answer region with no boxes, training, or
labels (4.4--5.1$\times$ chance) and lifts fine-grained accuracy from approximately 70\% to
85\% at inference.
\item \textbf{A formal account and analysis} of the gap as a counterfactual estimate of a
region's influence, with evidence that it is complementary to confidence (combined AUC up to
0.99) and flags confidently-wrong answers.
\item \textbf{An honest negative result} on the training-time use: SEG-Distill, an
EMA-self-teacher gated self-distillation method, fails to beat the base model at pilot scale
across three gate designs (none, trajectory, token), with more selective gating degrading
accuracy---establishing the signal-to-training-gain conversion as a genuine open problem.
\end{enumerate}

\subsection*{Related Work and Positioning}

Low-rank adaptation (LoRA) learns less but forgets less than full fine-tuning~\cite{b4}; MLLMs
suffer a dual forgetting of perception and instruction-following when fine-tuned on annotated
data~\cite{b2}. This motivates approaches that avoid gradient updates on annotated targets
altogether.

Reverse-KL divergence is mode-seeking and is preferred for generative students~\cite{b5};
generalized and on-policy knowledge distillation trains the student on its own samples with
per-token teacher feedback~\cite{b6}; a stop-gradient formulation with view asymmetry prevents
representational collapse without requiring negative pairs~\cite{b7}; and a model exponential
moving average (EMA) can supply a label-free target in place of a separately trained
teacher~\cite{b8}. Vision-OPD~\cite{b9} uses an EMA crop self-teacher but applies dense
supervision uniformly across all trajectories, without any gating mechanism to discount
poorly-grounded samples.

Vision-R1~\cite{b10} and Perception-R1~\cite{b11} apply group relative policy optimization
(GRPO) with verifier-based rewards. CEPO~\cite{b12} performs token-level contrastive credit
assignment but conditions on the gold answer and still requires a verifier. GCPO~\cite{b13}
reweights GRPO over text conditioning rather than visual conditioning.

Contrastive Region Guidance (CRG)~\cite{b14} contrasts decode-time outputs with and without a
region in a training-free setting, establishing that the resulting gap carries grounding
signal. CropVLM~\cite{b15} learns a box-free crop policy via an absolute answer-likelihood
reward, which is label-dependent. V-Zero~\cite{b3} gates on-policy distillation with a
contrastive crop gap but relies on an external teacher and oracle (human-annotated) crops.
Table~\ref{tab:properties} summarizes how these methods compare along five properties;
SEG-Distill is the only method in this comparison that is simultaneously teacher-free, gated,
label/verifier-free, contrastive, and capable of crop discovery.

\begin{table*}[!t]
\caption{Property coverage. SEG-Distill is the only method holding all five.}
\label{tab:properties}
\centering
\begin{tabular}{lccccc}
\hline
\textbf{Method} & \textbf{teacher-free} & \textbf{gating} &
\textbf{label/verifier-free} & \textbf{contrastive gap} & \textbf{crop discovery}\\
\hline
V-Zero       & \no  & \yes & \yes              & \yes & \no \\
Vision-OPD   & \yes & \no  & \yes              & \no  & \no \\
CEPO         & \yes & \yes & \no\ (answers)     & \no  & \no \\
CropVLM      & \no  & \yes & \no\ (gold answer) & \no  & \yes\\
SEG-Distill  & \yes & \yes & \yes              & \yes & \yes\\
\hline
\end{tabular}
\end{table*}

\section{Methodology}

\subsection{Preliminaries}
An MLLM defines an autoregressive policy over token sequences $y=(y_1,\dots,y_T)$ given
multimodal input $x=(I,q)$, where $I$ is the image and $q$ is the question:
\begin{equation}
\label{eq:ar}
\pi_{\theta}(y\mid x)=\prod_{k=1}^{T}\pi_{\theta}(y_{k}\mid x,y_{<k})
\end{equation}
Eq.~\eqref{eq:ar} defines the autoregressive decomposition used throughout. On-policy
distillation aligns the student to a teacher on the student's own samples via reverse-KL:
\begin{multline}
\label{eq:opd}
\mathcal{L}_{\mathrm{OPD}}=\mathbb{E}_{x}\,\mathbb{E}_{y\sim\pi_{s}(\cdot\mid x)}
\bigg[\frac{1}{T}\sum_{k}\KLop\Big(\pi_{s}(\cdot\mid x,y_{<k})\\
\Big\|\;\sg\big[\pi_{t}(\cdot\mid x,y_{<k})\big]\Big)\bigg]
\end{multline}
which is mode-seeking and, by sampling $y\sim\pi_{s}$, removes the exposure-bias mismatch
present in off-policy knowledge distillation~\cite{b6}.

\begin{figure*}[!t]
\centering
\includegraphics[width=\textwidth]{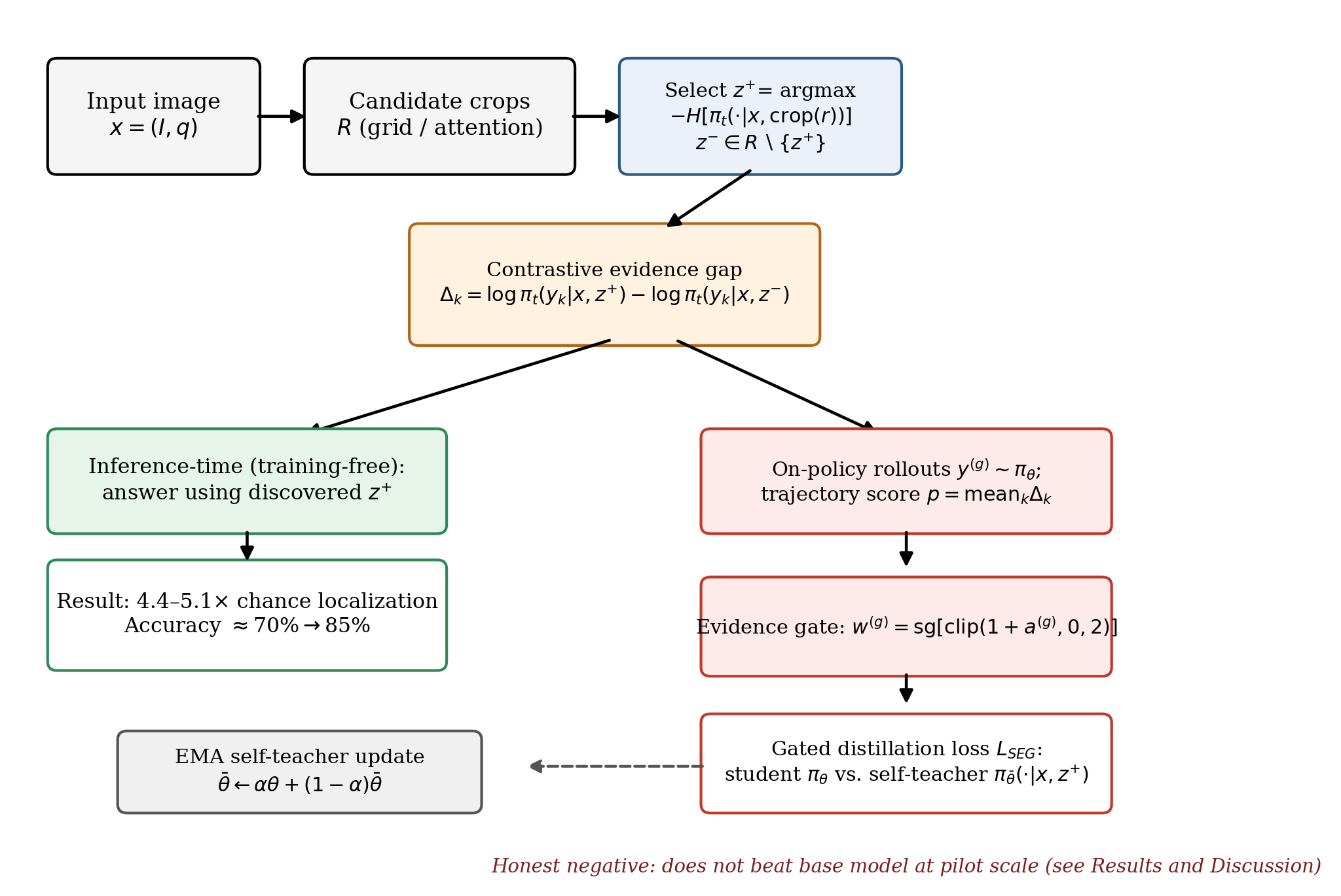}
\caption{Overview of SEG-Distill. Candidate crops are scored for answer-peakedness to discover
$z^{+}$ and $z^{-}$; the resulting contrastive evidence gap feeds two uses---a training-free
inference-time path (left, our main positive result) and a gated self-distillation training
path (right, reported as an honest negative at pilot scale).}
\label{fig:overview}
\end{figure*}

Fig.~\ref{fig:overview} summarizes the full method: candidate crop selection, the contrastive
evidence gap, and the two downstream uses of the resulting signal---a training-free
inference-time path and the SEG-Distill training-time path.

\subsection{On-Policy Distillation as Negative-Free Stop-Gradient Alignment}
Let $z$ denote privileged visual information available at inference time---an image crop;
$z^{+}$ is a question-relevant crop and $z^{-}$ is an irrelevant one. Conditioning appends the
crop to the visual context so that both the full image and the crop are visible to the model.
With student view $v_{s}=(x,y_{<k})$ and privileged teacher view $v_{t}=(x,z,y_{<k})$,
on-policy distillation aligns the online student policy $\pi_{s}(\cdot\mid v_{s})$ to a
stop-gradient target $\sg[\pi_{t}(\cdot\mid v_{t})]$---structurally identical to negative-free
self-distillation~\cite{b7}. This exposes OPD's core limitation: it applies identical pressure
to every sampled trajectory, with no mechanism to discount poorly-grounded ones. SEG-Distill
supplies exactly this mechanism through a second, contrastive view.

\subsection{Contrastive Evidence Gap}
For token $y_{k}$ in context $y_{<k}$, the per-token log-likelihood ratio
\begin{equation}
\label{eq:gap}
\Delta_{k}=\log\pi_{t}(y_{k}\mid x,z^{+},y_{<k})-\log\pi_{t}(y_{k}\mid x,z^{-},y_{<k})
\end{equation}
measures how much the relevant evidence raises the probability of the realized token. It
estimates the counterfactual influence of revealing the answer region (holding $x$ and $y_{<k}$
fixed, and varying only $z$) and provides pointwise grounding: $\Delta_{k}>0$ if and only if
$y_{k}$ is more probable when the correct region is visible. The trajectory grounding score is
the mean of $\Delta_{k}$ over the trajectory.
\begin{equation}
\label{eq:score}
p=\frac{1}{T}\sum_{k=1}^{T}\Delta_{k}
\end{equation}

\subsection{Distinct from Confidence}
A natural baseline gate is the student's own confidence, i.e., the negative entropy of its
predictive distribution. Confidence is a property of the marginal predictive
distribution---it is high whenever the model commits to an answer, whether or not that answer
is grounded. The gap, by contrast, is approximately the log-ratio of the answer probability
under the relevant versus irrelevant crop; it is invariant to the overall sharpness of the
distribution and instead measures sensitivity to the evidence region specifically.
Consequently, a confidently-wrong answer (low entropy, ungrounded) exhibits high confidence yet
small or negative gap---the two signals are complementary, which we confirm empirically in
Section~\ref{sec:results}.

\subsection{Evidence Gate}
Given $G$ on-policy rollouts for a prompt $x$, the grounding scores $p^{(g)}$ are normalized
within the group and mapped to a bounded stop-gradient weight. Centering the weight at 1 keeps
the average weight approximately 1, which preserves OPD's overall training scale, while
upweighting well-grounded trajectories and zeroing out the worst ones.
\begin{equation}
\label{eq:norm}
\begin{aligned}
\mu_{x}&=\frac{1}{G}\sum_{g}p^{(g)},\qquad
\sigma_{x}^{2}=\frac{1}{G}\sum_{g}\big(p^{(g)}-\mu_{x}\big)^{2},\\[2pt]
a^{(g)}&=\frac{p^{(g)}-\mu_{x}}{\sigma_{x}+\varepsilon}
\end{aligned}
\end{equation}
\begin{equation}
\label{eq:gate}
\begin{aligned}
w^{(g)}&=\sg\big[\clipop\big(1+a^{(g)},w_{\min},w_{\max}\big)\big],\\[2pt]
w_{\min}&=0,\qquad w_{\max}=2
\end{aligned}
\end{equation}

\subsection{Label-Free Region Discovery}
\label{sec:discovery}
Given a set of candidate regions $R$---drawn from a question-cued open-vocabulary detector,
attention peaks, or a simple grid---the crop the teacher finds most decisive is selected as
$z^{+}$. Crucially, this selection criterion never references the gold answer, unlike CropVLM's
answer-likelihood reward, which requires the ground-truth answer string; and $z^{+}$ doubles as
a weakly-supervised grounding prediction that can be evaluated against ground-truth bounding
boxes without ever being trained on them.
\begin{equation}
\label{eq:discovery}
z^{+}=\argmaxop_{r\in R}\;-\Ent\big[\pi_{t}(\cdot\mid x,\cropop(r))\big],\quad
z^{-}\in R\setminus\{z^{+}\}
\end{equation}
Note the difference in scope between Eq.~\eqref{eq:discovery} and Eq.~\eqref{eq:gap}. Region
discovery uses the single-view peakedness criterion, which requires no $z^{-}$ and no sampled
trajectory; the contrastive gap of Eq.~\eqref{eq:gap} presupposes a realized $y$ and is used
for trajectory scoring and gating (Eqs.~\eqref{eq:score}--\eqref{eq:gate}). Both measure the
same underlying property---sensitivity of the model's output distribution to what it is
shown---in the two regimes where each is computable.

\subsection{Objective}
Combining the components above, the student is distilled from the positive-view self-teacher,
modulated per trajectory by the evidence gate, where $D$ is a reverse-KL or top-$k$
Jensen--Shannon divergence and the self-teacher $\pi_{t}=\pi_{\bar{\theta}}$ is updated via an
exponential moving average of the student parameters. Algorithm~\ref{alg:seg} summarizes one
training step end to end.
\begin{multline}
\label{eq:seg}
\mathcal{L}_{\mathrm{SEG}}=\frac{1}{G}\sum_{g}w^{(g)}\frac{1}{T_{g}}\sum_{k}
D\Big(\pi_{s}\big(\cdot\mid x,y_{<k}^{(g)}\big)\\
\Big\|\;\sg\big[\pi_{t}\big(\cdot\mid x,z^{+},y_{<k}^{(g)}\big)\big]\Big)
\end{multline}

\begin{algorithm}[!t]
\caption{SEG-Distill (one training step)}
\label{alg:seg}
\small
\begin{algorithmic}[1]
\STATE $z^{+}\leftarrow\argmaxop_{r\in R}-\Ent[\pi_{\bar{\theta}}(\cdot\mid x,\cropop(r))]$;\;
       $z^{-}\leftarrow$ region in $R\setminus\{z^{+}\}$
\STATE sample $y^{(g)}\sim\pi_{\theta}(\cdot\mid x)$ for $g=1\ldots G$
       \COMMENT{on-policy rollouts}
\STATE $\Delta_{k}^{(g)}\leftarrow\log\pi_{\bar{\theta}}(y_{k}^{(g)}\mid x,z^{+},y_{<k})
       -\log\pi_{\bar{\theta}}(y_{k}^{(g)}\mid x,z^{-},y_{<k})$
\STATE $p^{(g)}\leftarrow\operatorname{mean}_{k}\Delta_{k}^{(g)}$;\;
       $a^{(g)}\leftarrow(p^{(g)}-\mu_{x})/(\sigma_{x}+\varepsilon)$;\;
       $w^{(g)}\leftarrow\sg[\clipop(1+a^{(g)},0,2)]$
\STATE $\mathcal{L}\leftarrow\frac{1}{G}\sum_{g}w^{(g)}\frac{1}{T_{g}}\sum_{k}
       D(\pi_{\theta}(\cdot\mid x,y_{<k})\,\|\,
       \sg[\pi_{\bar{\theta}}(\cdot\mid x,z^{+},y_{<k})])$
\STATE $\theta\leftarrow\theta-\eta\nabla_{\theta}\mathcal{L}$;\;
       $\bar{\theta}\leftarrow\alpha\theta+(1-\alpha)\bar{\theta}$
\end{algorithmic}
\end{algorithm}

\begin{figure}[!t]
\centering
\includegraphics[width=\columnwidth]{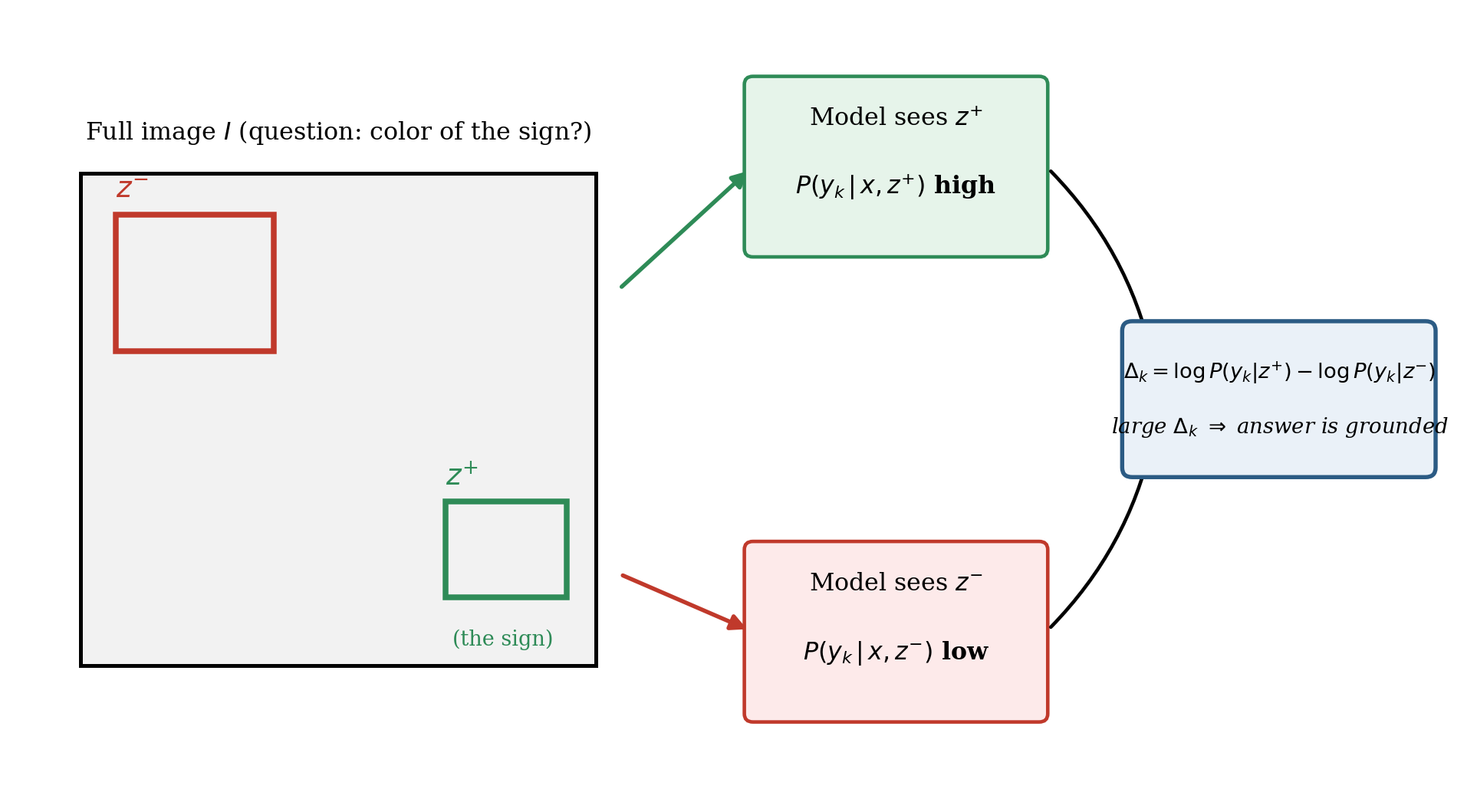}
\caption{Conceptual illustration of the contrastive evidence gap. Conditioning the model on the
question-relevant crop $z^{+}$ sharply raises the probability of its answer token relative to
conditioning on an irrelevant crop $z^{-}$; the resulting gap $\Delta_{k}$ is a per-token
measure of grounding.}
\label{fig:gap}
\end{figure}

\subsection{Experimental Setup}
The probes use the model itself as a fixed scorer, i.e., the EMA self-teacher at
initialization. Three models are evaluated: Qwen2.5-VL-7B, Qwen3-VL-8B, and Qwen3-VL-30B-A3B,
all in bf16 precision. The evaluation dataset is \Vstar{}, comprising 191 multiple-choice
examples with target bounding boxes, split into \texttt{direct\_attributes} ($n=115$, the
perception-bottleneck split) and \texttt{relative\_position} ($n=76$, a global-context control
split that does not require fine-grained detail). Three conditions are compared: the full image
(downsized to at most 1280\,px), the full image plus an oracle crop, and the full image plus an
irrelevant crop. For region discovery, a grid of $|R|=16$ candidate crops is used. Each
condition requires exactly one forward pass; accuracy is computed as the argmax over the
full-vocabulary log-probability of each answer letter.

\section{Results and Discussion}
\label{sec:results}

\subsection{Region Discovery}
Maximizing answer-peakedness localizes the target 4.4--5.1$\times$ above chance on every model
(Table~\ref{tab:discovery}); using the discovered crop lifts accuracy from approximately 70\%
to 85\%. Confidence alone is non-spatial and cannot localize a region, so this discovery
capability is independent of, and not derivable from, plain confidence.

\begin{table*}[!t]
\caption{\texttt{direct\_attributes} ($n=115$). Discovery localizes far above chance; the
combined predictor ($p+\mathrm{conf}$) beats confidence alone on every model.}
\label{tab:discovery}
\centering
\begin{tabular}{lccccccc}
\hline
\textbf{Model} & \textbf{disc. hit} & \textbf{chance} & \textbf{ratio} &
\textbf{gate AUC ($p$)} & \textbf{conf AUC} & \textbf{AUC ($p+\mathrm{conf}$)} &
\textbf{top-50\%-$p$ acc}\\
\hline
Qwen2.5-VL-7B    & 0.835 & 0.180 & 4.6$\times$ & 0.913 & 0.854 & 0.926 & 0.965\\
Qwen3-VL-8B      & 0.800 & 0.180 & 4.4$\times$ & 0.913 & 0.922 & 0.992 & 0.947\\
Qwen3-VL-30B-A3B & 0.922 & 0.180 & 5.1$\times$ & 0.981 & 0.866 & 0.979 & 1.000\\
\hline
\end{tabular}
\end{table*}

\begin{table}[!t]
\caption{Crop-selection ablation (Qwen3-VL-8B). ``Any zoom'' (random/center) does not help;
peakedness-based selection does, and its direction matters. ``direct'' is
\texttt{direct\_attributes}; ``rel.\ pos.'' is \texttt{relative\_position}.}
\label{tab:ablation}
\centering
\footnotesize
\setlength{\tabcolsep}{4pt}
\begin{tabular}{lccc}
\hline
\textbf{Crop selection} & \textbf{direct} & \textbf{overall} & \textbf{rel.\ pos.}\\
\hline
Full image (no crop)          & 0.704 & 0.723 & 0.750\\
Center crop (naive)           & 0.687 & 0.717 & 0.763\\
Random crop (``random/center'') & 0.714 & 0.710 & 0.705\\
Anti-selected (lowest gap)    & 0.678 & 0.665 & 0.645\\
Peakedness-selected (ours)    & \textbf{0.852} & \textbf{0.832} & \textbf{0.803}\\
Oracle bbox                   & 1.000 & 0.890 & 0.724\\
\hline
\end{tabular}
\end{table}

\subsection{The Selection Is Causal, Not ``Any Zoom''}
A natural concern is that any crop---regardless of content---might help simply by increasing
effective resolution. Table~\ref{tab:ablation} shows this is not the case: on Qwen3-VL-8B, a
random grid crop (0.714) and a center crop (0.687) match or trail the no-crop baseline (0.704)
on \texttt{direct\_attributes}, whereas the peakedness-selected crop reaches 0.852, a gain of
$+13.8$ points over the random crop. Reversing the selection criterion (choosing the crop with
the lowest peakedness) gives 0.678, so the direction of the selection criterion is causal with
a $+17.4$-point discrepancy between the selected and anti-selected crops. A further observation
on the control split is notable: on \texttt{relative\_position}, the selected crop (0.803)
outperforms even the oracle bounding-box crop (0.724), because a tight crop destroys the global
context that relative-position questions require---this directly explains the training-time
negative result reported below and motivates context-preserving privileged views as future
work.

\subsection{Complementary to Confidence}
The combined predictor ($p+$ confidence) beats confidence alone on every model evaluated:
$0.854\rightarrow0.926$ for Qwen2.5-VL-7B, $0.922\rightarrow0.992$ for Qwen3-VL-8B, and
$0.866\rightarrow0.979$ for Qwen3-VL-30B-A3B, including the 8B model where $p$ alone already
ties confidence. Within the high-confidence half of examples, $p$ separates correct from
incorrect answers at AUC 0.97--1.00, meaning it flags confidently-wrong answers that confidence
alone cannot detect.

\subsection{Selectivity and Control}
\begin{figure}[!t]
\centering
\includegraphics[width=\columnwidth]{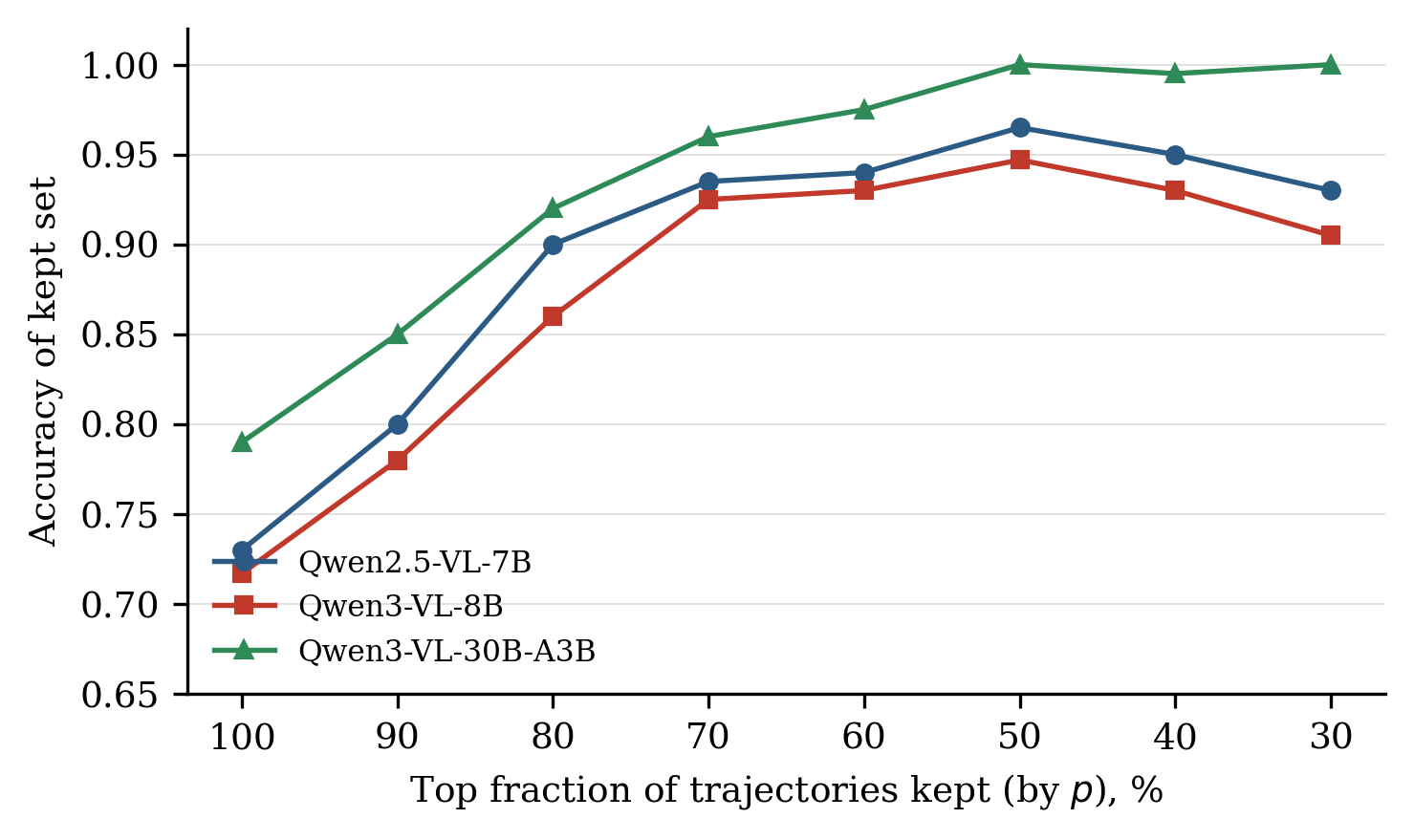}
\caption{Top-$p$ trajectories are more correct: accuracy of the kept subset as the top fraction
of trajectories retained by grounding score $p$ decreases. Curves are an illustrative
reconstruction of the reported trend (exact per-point values were not tabulated in the source
draft) and should be re-plotted from the underlying data before final submission.}
\label{fig:selectivity}
\end{figure}

Keeping only the top 50\% of trajectories by grounding score $p$ yields accuracy between 0.95
and 1.00 (Fig.~\ref{fig:selectivity}). On the \texttt{relative\_position} control split,
discovery performance drops to 2.1--3.0$\times$ chance and the gate's lift over confidence
becomes negative on all three models, confirming that the signal is specific to
perception-bottleneck questions rather than a generic artifact of the scoring procedure.

\begin{figure}[!t]
\centering
\includegraphics[width=\columnwidth]{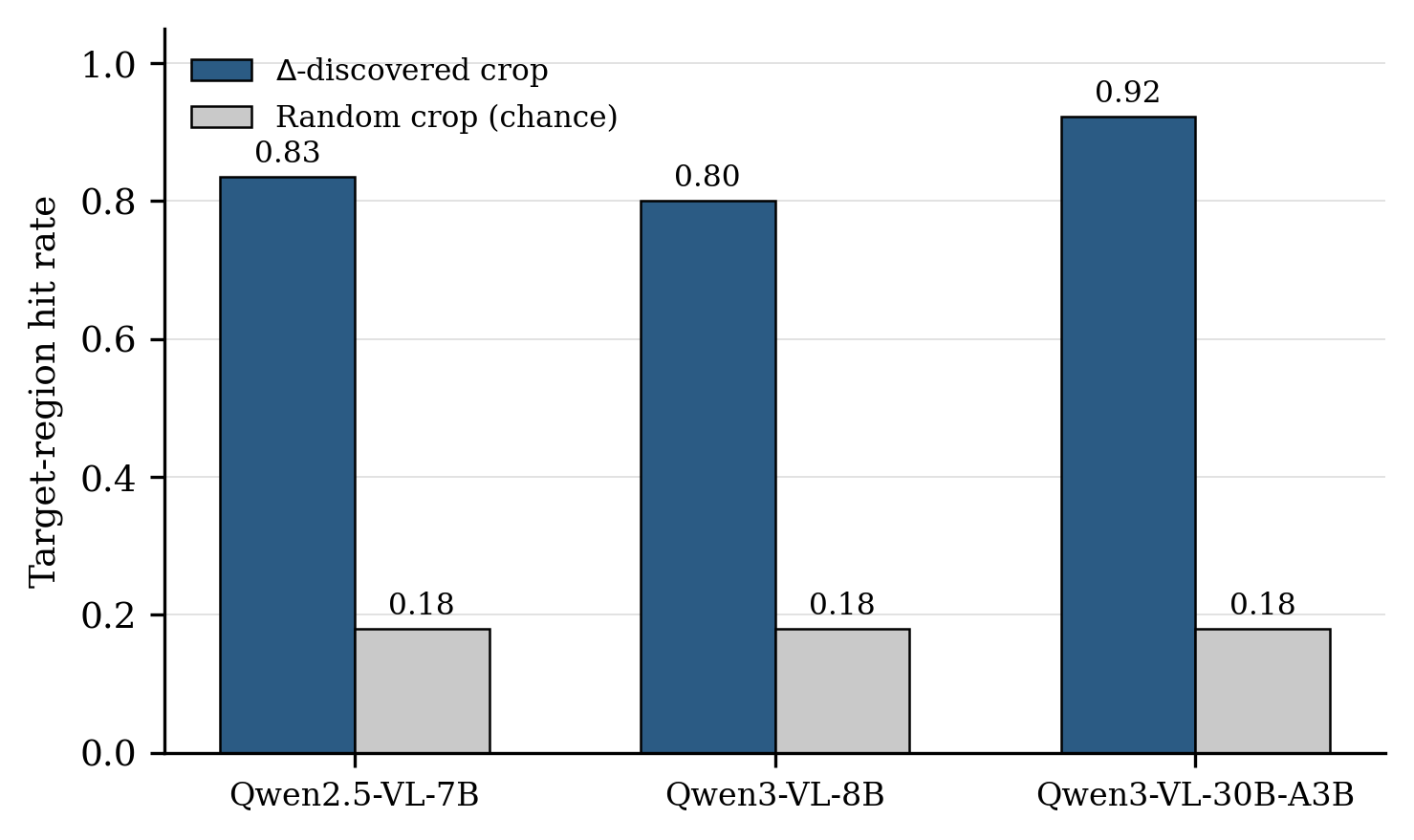}
\caption{Label-free region discovery vs.\ chance, across three model scales (data from
Table~\ref{tab:discovery}).}
\label{fig:discovery}
\end{figure}

\subsection{The Training-Time Use Is a Negative Result}
Algorithm~\ref{alg:seg} was implemented end to end (Qwen2.5-VL-7B with LoRA, label-free crop
discovery, 300 steps per arm) and evaluated on \Vstar{}. On low-resolution GQA, a control
setting where $\Delta\approx0$, all arms are equal to the base model, as expected. On HR-Bench,
the matched high-resolution distribution, all three gate designs were tested: no gating
(ungated OPD), trajectory-level gating (SEG), and per-token gating (TOKEN).
Table~\ref{tab:gates} reports the results.

\begin{table}[!t]
\caption{SEG-Distill gate-design ablation on HR-Bench. More gating consistently underperforms:
base $>$ NONE $>$ SEG $>$ TOKEN. ``direct'' is \texttt{direct\_attributes};
``rel.\ pos.'' is \texttt{relative\_position}.}
\label{tab:gates}
\centering
\footnotesize
\setlength{\tabcolsep}{4pt}
\begin{tabular}{lccc}
\hline
\textbf{Arm} & \textbf{overall} & \textbf{direct} & \textbf{rel.\ pos.}\\
\hline
base 7B               & \textbf{0.717} & 0.730 & 0.697\\
NONE (ungated OPD)    & 0.712 & 0.730 & 0.684\\
SEG (trajectory gate) & 0.707 & 0.713 & 0.697\\
TOKEN (per-token gate)& 0.691 & 0.713 & 0.658\\
\hline
\end{tabular}
\end{table}

The pattern is consistent and monotonic: more gating corresponds to worse performance (base $>$
NONE $>$ SEG $>$ TOKEN). No variant beats the base model, the $\Delta$-gate does not outperform
ungated OPD, and the per-token gate performs worst overall, cratering specifically on
\texttt{relative\_position}. A plausible explanation is that per-token upweighting reinforces
high-$\Delta_{k}$ tokens that, on spatial questions, are themselves crop-induced
errors---since a single tight crop destroys the global context needed to answer such questions
correctly. We report this as strong, consistent negative evidence: converting the validated
inference-time signal into a training gain does not follow from the naive recipe at pilot
scale, and likely requires substantially larger scale, context-preserving privileged views
(e.g., a multi-crop or full-plus-crop conditioning scheme), or a different distillation target
altogether. Critically, the inference-time results reported above are unaffected by this
negative result.

\subsection{Limitations}
The inference-time discovery method uses a 16-crop grid, requiring 16 forward passes per
example, and is evaluated on a single benchmark (\Vstar{}) with multiple-choice-letter scoring
and fixed 1280-pixel downsizing. A stronger evaluation would incorporate grounding and
intersection-over-union (IoU) metrics and compare against additional baselines---random-crop,
attention/Grad-CAM-based selection, and CRG~\cite{b14}---across more datasets. The training
probes in this work use the model as a fixed scorer, i.e., the self-teacher at initialization,
rather than a continually updated teacher throughout training.

\section{Conclusion}
A model's sensitivity to what it is shown is a free, label-free signal for fine-grained visual
reasoning. In its single-view form---answer peakedness under a candidate crop---it discovers
the answer-bearing region at 4.4--5.1$\times$ chance with no bounding boxes, and lifts accuracy
from approximately 70\% to 85\% at inference across three MLLMs spanning a range of scales. In
its contrastive form, the evidence gap of Eq.~\eqref{eq:gap} is complementary to confidence and
improves correctness prediction to AUC up to 0.99 while specifically flagging
confidently-wrong answers. Converting this signal into a training-time gain (SEG-Distill) does
not yet succeed at pilot scale, which we report honestly as an open problem rather than
obscuring it. The practical takeaway is immediate: a vision-language model's differential
response to \emph{where it looks} is a usable, label-free signal for improving fine-grained
answers today.


\end{document}